\documentclass[review]{elsarticle}

\usepackage{algorithm}
\usepackage{algorithmic}
\usepackage{caption}
\usepackage{textcomp}
\usepackage{amsmath, epsfig}
\usepackage{amssymb}
\usepackage{amsfonts}
\usepackage{latexsym}
\usepackage{tikz}
\usepackage{array}
\usepackage {subcaption}

\usepackage{lineno,hyperref}
\modulolinenumbers[10]

\begin{document}

\begin{frontmatter}

%
%
%
%

\title{Label-Removed Generative Adversarial Networks Incorporating with K-Means}
\author[add1]{Ce~Wang}
\author[add2]{Zhangling~Chen\corref{cor1}}
\ead{zhanglingchen0214@tju.edu.cn}
\author[add3]{Kun~Shang}

\cortext[cor1]{Corresponding author.}

\address[add1]{Center for Combinatorics, Nankai University, Tianjin 300071, P.R. China}
\address[add2]{Center for Applied Mathematics, Tianjin University, Tianjin 300072, P.R. China}
\address[add3]{College of Mathematics and Econometrics, Hunan University, Changsha, Hunan 410082, P.R. China}

\begin{abstract}
Generative Adversarial Networks (GANs) have achieved great success in generating realistic images.
Most of these are conditional models, although acquisition of class labels
is expensive and time-consuming in practice. To reduce the dependence on labeled data,
we propose an un-conditional generative adversarial model, called K-Means-GAN (KM-GAN), which
incorporates the idea of updating centers in K-Means into GANs. Specifically, we redesign the framework of
GANs by applying K-Means on the features extracted from the discriminator. With obtained labels from K-Means,
we propose new objective functions from the perspective of deep metric learning (DML).
Distinct from previous works, the discriminator is treated as a feature extractor rather than a classifier in KM-GAN,
meanwhile utilization of K-Means makes features of the discriminator more representative. Experiments are conducted
on various datasets, such as MNIST, Fashion-10, CIFAR-10 and CelebA, and show that the quality of
samples generated by KM-GAN is comparable to some conditional generative adversarial models.

\end{abstract}


\end{frontmatter}

\section{Introduction}

Generative models have been an active but challenging research field in traditional machine learning because of
the intractability of many probabilistic computations arising in approximating maximum likelihood estimation (MLE).
To avoid these computations, Generative Adversarial Network (GAN) \cite{goodfellow2014generative} greatly improves the
quality of generated images by implicitly modeling the target distribution via neural
networks instead of approximation of intractable likelihood functions in capturing data distribution.
To better utilize the information about data structure in labeled data, Conditional GAN (CGAN) \cite{mirza2014conditional}
feeds real labels along with images and generate more realistic images.
Unfortunately, CGAN and subsequent extensions \cite{dai2017metric, radford2015unsupervised,
dou2017metric, mao2017least, huang2017stacked} suffer from a challenge that they require large
amounts of labeled data which is expensive or even impossible to acquire in practice.

To decrease the dependence of GANs on labeled data, it would be nicer
to find a substitution to replace the role of real labels. It is well known that representation learning enables machine learning
models to get more information about data structure and class distribution. A commonly and
widely used method in representation learning is to employ K-Means.
Recent works \cite{xie2016unsupervised, yang2017towards, yang2016joint, aljalbout2018clustering}
have improved clustering results through jointly training K-Means and deep neural networks.
By fusing K-Means with the powerful nonlinear
expressiveness of neural networks, they get ``K-Means-friendly" \cite{yang2017towards} representations, i.e.,
features that are more representative for clustering tasks.
But most of these neural networks are realized by a
pre-trained auto-encoder on large-scale datasets like ImageNet, which means they still
utilize prior knowledge (real-label) supervision.

Inspired by the success of jointly training of neural networks and K-Means on clustering tasks,
Variational deep embedding (VaDE) \cite{jiang2017variational} and Joint Generative MomentMatching
Network (JGMMN) \cite{gao2018joint} instead combine generative models
with clustering methods and achieve competitive results not only on clustering, but also on generating tasks.
More specifically, VaDE proposes continuous clustering objectives for Variational Autoencoder (VAE) \cite{kingma2013auto}
and JGMMN augments original loss functions of Generative Moment Matching Networks (GMMN) \cite{li2015generative}
with regularization terms to constrain latent variables.
On the other hand, authors of \cite{premachandran2016unsupervised} perform
K-Means on features of the top layer of discriminators in GAN and Info-GAN \cite{chen2016infogan}
respectively and  show that features of Info-GAN are obviously more ``K-Means-friendly" than regular GAN.
This implies that constrains on the latent space of GANs induce more representative features.
Furthermore, extensions of GANs \cite{ben2018gaussian, mukherjee2018clustergan} give
state-of-the-art results on clustering by fusing GANs with clustering methods.
Although these works have achieved exciting results on clustering results by combining advantages
of GANs and clustering method, utilizing clustering methods
to improve the quality of generating images of GANs also deserves more attentions.
This brings the main motivation of our work:
\textbf{Can we re-design the framework of GANs in an un-conditional manner and utilize the capability of K-Means
on representation learning to replace the role of real labels?}

In order to make use of clustering labels of K-Means to direct the generating process as real labels in GANs,
we consider operating K-Means on the top layer of the discriminator.
But the main difficulty is how to deal
with the un-differentiable objective of K-Means using Stochastic Gradient Decent (SGD) \cite{kingma2014adam}.
Deep Embedded Clustering (DEC) \cite{xie2016unsupervised} straightforwardly separates the optimization into
updating centers and network parameters successively. Another CNN-based method \cite{hsu2018cnn} also adopts
this technique and further proposes a feature drift compensation scheme to mitigate the drift error caused
different optimization direction of K-Means and regular
loss functions. Then Deep Clustering Network (DCN) \cite{yang2017towards} introduces a defined ``pretext" objective,
a mathematical combination of reconstruction loss and K-Means clustering objective, and
optimize K-means with back-propagation. Quite recently, Deep K-Means \cite{fard2018deep} proposes a continuous reparametrization
of the objective of K-Means clustering to optimize it with SGD.

Motivated by these works, we propose an un-conditional generative adversarial model,
named K-Means-GAN (KM-GAN), which embeds the idea of updating centroids of K-means into the framework of GANs.
As the illustration of the framework of our model in Fig. \ref{flow-chart}, it conducts the discriminator as
a nonlinear feature extractor and utilizes K-Means clustering
algorithm for getting more representative features. Further, we employ obtained results of K-Means instead of one-hot
real labels to direct the generator in the generating process. Then we propose objectives containing
clustering labels from the perspective of deep metric learning (DML) to let the optimization direction of K-Means agree with
the generating process. The specific optimization process includes three terms to alternately
optimize, of which the ``center-loss" term tries to
pull the corresponding centers of real and generated images closer.
Furthermore, the objective of the discriminator is to minimize the distance between real samples
and their corresponding centers and maximize the distance between fake samples and their corresponding real centers.
Meanwhile, the loss function of the generator, which is
interpreted as an adversarial term, attempts to approximate
the target distribution by decreasing the distance between generated samples and their corresponding real centers.

\makeatletter\def\@captype{figure}\makeatother
\begin{center}
\includegraphics[height=4.8cm]{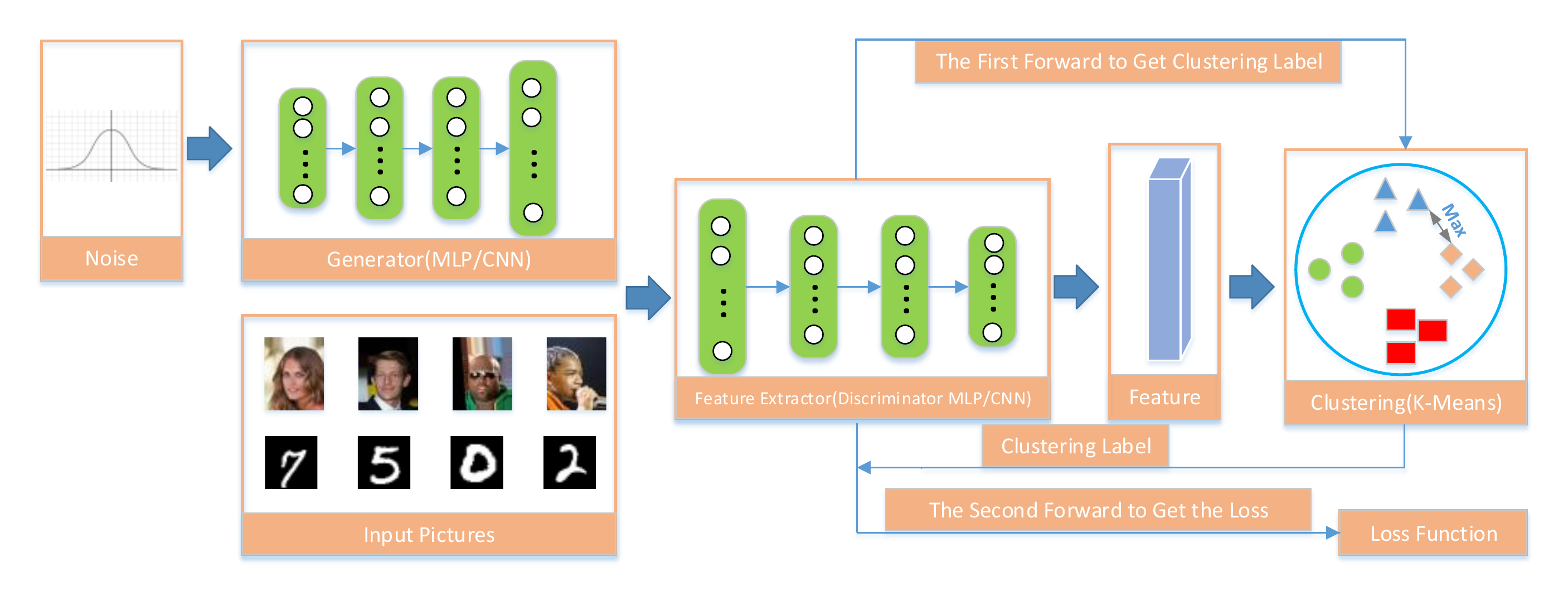}
\caption{The framework of KM-GAN. Notice that the first forward pass
is to get the clustering labels. In the second forward pass, the clustering labels
and data are feeded to back-propagate the obtained loss functions.}
\label{flow-chart}
\end{center}

\textbf{Contribution} To the best of our knowledge, our work is the first to
attempt to combine training unsupervised K-Means algorithm with GAN model simultaneously
through SGD for generating tasks. Our main contributions are summarized as follows:

\begin{itemize}
  \item We propose an un-conditional implementation of GANs, called K-Means-GAN (KM-GAN), and equip it
  with new objective functions from the perspective of DML.

  \item We incorporate GANs with the idea of traditional K-Means and utilize obtained labels,
  replacing the role of real labels, to direct the generating process and get more representative features.

  \item We empirically show that KM-GAN is capable to generate diverse samples and the quality of generated images on several real datasets is competitive to that of conditional GANs.
\end{itemize}

\section{Background}

In this section, we introduce notations and briefly review preliminary knowledge,
including the framework of GANs and K-Means.
The notations provided in this section will also be used in subsequent sections.

\subsection{Notations}
Throughout the paper, we use $b$ for the batch size, $D$ for the discriminator, $G$ for
the generator and $k$ for the pre-defined number of classes.

\subsection{Framework of GANs}
GAN \cite{goodfellow2014generative} consists of two components: a discriminator $D$ and a generator $G$ which
are both realized by the neural networks.
The main idea is actually an adversarial training procedure between them. Throughout the adversarial training,
the generator $G$ maps samples from a prior noise distribution, such as gaussian distribution,
to the data space while the discriminator $D$ estimates the probability that its inputs coming from
real data distribution rather than generated distribution.

More specifically, given a noise distribution ${P}_{\mathbf{z}}$ and training samples $\mathbf{x}\sim
{P}_{\mathbf{x}}$, the adversarial training contains two steps.
Firstly, the generator maps noises $\mathbf{z}$ from ${P}_{\mathbf{z}}$ to $G(\mathbf{z})$ and update
parameters of the discriminator while fix parameters of the generator by optimizing the objective of $D$ as follows:
\begin{equation}\label{discriminator of gan}
  \min_{D} \mathbb{E}_{\mathbf{x}\sim{p}_{\mathbf{x}}}[ \log D(\mathbf{x})]+\mathbb{E}_{\mathbf{z}\sim{p}_{\mathbf{z}}}[\log (1-D(G(\mathbf{z})))].
\end{equation}

Then fix parameters of $D$ and update parameters of $G$ to approximate target distribution by optimizing
the loss function of $G$ as follows:
\begin{equation}\label{generator of gan}
  \max_{G}\mathbb{E}_{\mathbf{z}\sim{p}_{\mathbf{z}}}[\log (1-D(G(\mathbf{z})))].
\end{equation}

In order to generate more realistic images, CGANs \cite{mirza2014conditional} implements GANs
with one-hot real labels, which provides supplementary information of class distribution for generating process.
This method qualitatively and quantitatively improve the performance in generating tasks.
Recent works \cite{doersch2016tutorial, ren2016conditional}
then extend VAE and GMMN based on this technique for more realistic images.
Furthermore, Deep Convolutional GAN (DCGAN) \cite{radford2015unsupervised} designs a stable architecture utilizing convolutional
neural networks and raises several tricks to stabilize the adversarial training process.
On the other hand, lots of works \cite{dou2017metric, dai2017metric, arjovsky2017wasserstein, nowozin2016f, qi2017loss}
propose objectives for GANs to improve stability and image quality.

\subsection{K-Means}
K-Means \cite{macqueen1967some} is a traditional clustering method used to group a set of given data points
$\{\mathbf{{x}}_{i}\}_{i=1,2,\ldots,N}\in {\mathbb{R}}^{m}$ into $k$ clusters, where
$k$ is a pre-defined number. After randomly choosing $k$ points of data samples as initialized center,
the main algorithm is composed of two steps. The first is to assign clustering labels to each point
according to the Euclidean distance between the point and current the $k$ centers.
Then update new centers as the weighted average of points in each class.
The algorithm stops until each center do no change.
Formally, the cost function is as follows:
\begin{align}\label{objective of kmeans}
\begin{aligned}
  \min_{\mathbf{M}\in \mathbb{R}^{m*k}, {\mathbf{s}}_{i}\in\mathbb{R}^{k}} &
  \sum^{N}_{i=1}\| {\mathbf{x}}_{i} -\mathbf{M} {\mathbf{s}}_{i}\|^{2}_{2} \\
  \vspace{5pt}
  s.t.\quad & {s}_{ij}\in\{ 0,1\}, \mathbf{1}^{T}{\mathbf{s}}_{i}=1, \forall i, j,
\end{aligned}
\end{align}
where ${\mathbf{s}}_{i}$ is the one-hot clustering label of data point ${\mathbf{x}}_{i}$,
${s}_{ij}$ denotes the $j$th element of vector ${\mathbf{s}}_{i}$ and
$\mathbf{M}$ is a matrix, whose $k$ columns correspond to the $k$ centers.

As we can see from this formula, the performance of K-Means depends on both features and
initialized centers. So K-Means++ \cite{arthur2007k} is proposed to initialize centers with
a better procedure. Extensions \cite{hsu2018cnn, fard2018deep, premachandran2016unsupervised}
adopt the procedure and achieve surprising results. Then to deal with large-scale datasets and online scenarios,
Minibatch K-Means \cite{sculley2010web} proposes to use a batch of samples to update centers in each iteration.

\section{Proposed Method}

As mentioned before, we consider re-designing the framework of GANs and utilize results of K-means to
replace the role of one-hot real labels in a un-conditional manner. So we treat the discriminator
as a feature extractor instead of a classifier and operate K-Means on extracted features to produce
clustering labels which are viewed as substitution of real labels. With the obtained features and labels, we
propose our objectives from the perspective of DML to carry out adversarial learning.
More importantly, we come up with a ``center-loss" term to connect the optimization of adversarial learning
and centers updating in K-Means. In the following subsections, we first introduce proposed objectives and
optimization procedure in regular K-Means-GAN (KM-GAN). Then generalize it with regularization terms in
order to deal with more general datasets.

\subsection{Regular KM-GAN}
We first introduce the ``center-loss" term since it fills the gap between two different
optimization procedures of adversarial learning and K-Means,
which is important for the whole algorithm to work effectively.
The term is interpreted as a role to decrease the distance between corresponding
centers of real and generated images. Formally, the formula is as follows:
\begin{equation}\label{center-loss-constraint}
\begin{split}
  \min_{D,G} &\quad\quad{L}_{center} =\| \sum_{m=1}^{k} \frac{{\mathbf{c}}_{m}+\sum_{j=1}^{{j}_{{c}_{m}}}D({\mathbf{x}}_{{n}_{j, {c}_{m}}})}{1+{j}_{{c}_{m}}} -
  \sum_{m=1}^{k} \frac{{\widehat{\mathbf{c}}}_{m}+\sum_{j=1}^{{j}_{{\widehat{c}}_{m}}}D(G({\mathbf{z}}_
  {{n}_{j, {\widehat{c}}_{m} }}))}
  {1+{j}_{{\widehat{c}}_{m}}} \|_1 \\
  \vspace{5pt}
  s.t.       &\quad\quad {L}_{center} \geq {d}_{round},
\end{split}
\end{equation}
where $k$ is the pre-defined number of classes,
${\mathbf{c}}_{m}$ (${\widehat{\mathbf{c}}}_{m}$) is the $m$-th center
of features of real data (generated data) updated after last iteration, ${j}_{{c}_{m}} $ (${j}_{{\widehat{c}}_{m}}$)
is the number of features belonging to the
center ${\mathbf{c}}_{m}$ (${\widehat{\mathbf{c}}}_{m}$),
${n}_{j, {c}_{m}}$ (${n}_{j, {\widehat{c}}_{m}}$) denotes the position of corresponding feature of
real data (generated data) that is in class $m$ according to results of K-Means in the first forward pass
and ${d}_{round}$ is a hyperparameter needed to tune according to different datasets to avoid degeneration.

Indeed, ${L}_{center}$ calculates the difference of second order statistical magnitude, i.e.,
the average of $k$ centers, between features of real and generated images. The intension is to
keep centers of synthesized data not far away from that of real data and accelerate distribution approximation.
The exploration of minimizing statistical magnitudes is motivated by improvements of recent works \cite{wen2016discriminative, dou2017metric, li2015generative} on classification and generation tasks.
Especially, GMMN successfully approximates data distribution through minimizing
all orders of statistics, which is realized by the Gaussian kernel. So we intuitively
utilize the second order statistics and reuse the results of K-Means to propose the continuous term.
Experiments further show that KM-GAN fails to generate meaningful images even on MNIST without ``center-loss" term.

Although the ``center-loss" term is proposed to approximate the target distribution, we
still need objective functions for the discriminator and the generator to
finish the regular adversarial training. Firstly, we define the objective function of discriminator
as follows:
\begin{equation}\label{D-loss}
  \min_{D}\quad {L}_{D}  = \|D(\mathbf{x}) - C_{real} \|_2 - \|D(G(\mathbf{z})) - C_{gen} \|_2,
\end{equation}
where $C_{real}$ ($C_{gen}$), computed based on real centers, consists of $b$ center pieces for the
pre-defined batch size $b$. Each of these center pieces is the centroid of the real class that the feature
piece in the corresponding position of this batch belongs to.
It's natural to see that ${L}_{D}$ penalizes the distance between
each class of real data and their corresponding $k$ centers. The interpretation is to minimize intra-class
distance of each class in the feature space of real data from the viewpoint of DML.
On the contrary, ${L}_{D}$ maximizes the distance between generated data and centers of their corresponding
real classes to discriminate the counterfeit from real data.

On the other hand, the corresponding objective function of the generator is defined as follows:
\begin{equation}\label{G-loss}
  \min_{G} \quad {L}_{G}  = \|D(G(\mathbf{z})) - C_{gen} \|_2.
\end{equation}

Obviously, the effect of the objective is to compete with the discriminator to approximate the target distribution.
When decreasing the distance between synthesized data and centers of their corresponding $k$ real classes, the
features of generated images are distributed around each real center like features of real data. Then with the impact
of ``center-loss" to pull centers of real and fake data close, fake data distribution would approximate
target distribution finally. The term also plays a role as an adversarial term in the framework of KM-GAN.

\begin{algorithm}
\caption{Training algorithm for regular KM-GAN}
\label{alg1}
\begin{algorithmic}
\REQUIRE Real images $\mathbf{X}$, noise distribution${P}_{\mathbf{Z}}$,
pre-defined number of classes $k$, number of iterations $T$, batch size $b$
and hyper-parameters of Adam $\alpha, {\beta}_{1}, {\beta}_{2}$
\ENSURE Generated samples $G(\mathbf{z})$

\STATE Initialize parameters of $D$ and $G$ networks
\STATE Initialize $k$ real centers of mapped features $D(\mathbf{X})$ by K-Means++
\FOR{$t=1:T$}
\STATE Sample a batch $\{{\mathbf{x}}_{i}\}_{i=1}^{b}$ from real data $\mathbf{X}$
\STATE Sample a batch $\{{\mathbf{z}}_{i}\}_{i=1}^{b}$ from noise distribution ${P}_{\mathbf{Z}}$
\STATE Obtain features $\{D({\mathbf{x}}_{i})\}_{i=1}^{b}$ and $\{D(G({\mathbf{z}}_{i}))\}_{i=1}^{b}$
\STATE Obtain clustering labels according to Euclidean distance with current $k$ centers
\vspace{2pt}
\STATE ${grad}_{{\theta}_{d}}= {\triangledown}_{{\theta}_{d}}{L}_{D}$
\vspace{2pt}
\STATE ${\theta}_{d}=$Adam(${grad}_{{\theta}_{d}}, {\theta}_{d}, \alpha, {\beta}_{1}, {\beta}_{2}$)
\vspace{2pt}
\STATE ${grad}_{{\theta}_{g}}= {\triangledown}_{{\theta}_{g}}{L}_{G}$
\vspace{2pt}
\STATE ${\theta}_{g}=$Adam(${grad}_{{\theta}_{g}}, {\theta}_{g}, \alpha, {\beta}_{1}, {\beta}_{2}$)
\vspace{2pt}
\STATE ${grad}_{{\theta}_{d}, {\theta}_{g}}= {\triangledown}_{{\theta}_{d},
{\theta}_{g}}{L}_{center}$
\vspace{2pt}
\STATE ${\theta}_{d}, {\theta}_{g}=$Adam(${grad}_{{\theta}_{d}, {\theta}_{g}}, {\theta}_{d}, {\theta}_{g}, \alpha, {\beta}_{1}, {\beta}_{2}$)
\vspace{2pt}
\STATE Update centers via K-Means objective in Equation \ref{objective of kmeans}
\ENDFOR
\end{algorithmic}
\end{algorithm}

\subsection{Three-Step Alternating Optimization}
Optimizing network parameters of GANs and updating centers step by step is straightforward as in
DEC \cite{xie2016unsupervised}. But the different directions of these two steps make the optimization
more difficult. To deal with this issue, we utilize ``center-loss" term to bridge the gap. Especially,
the ``center-loss" term reuses results of K-Means and obtained features from the discriminator, which
builds a connection between these two steps. In the specific optimization, we first solve the subproblem
of adversarial learning, i.e., updating parameters of the discriminator and generator, respectively.
Then inspired by alternating optimization in \cite{yang2017towards}, we utilize ``center-loss" to re-update
parameters of $D$ and $G$ via SGD. With current parameters, we obtain centers in feature space at last by
Equation \ref{objective of kmeans}. The concretely three-step alternating optimization procedure is shown in
Algorithm \ref{alg1}.

In the described algorithm, we conduct K-Means++ technique to better initialize centers of features.
In addition, since the optimization of network parameters employs Adam \cite{kingma2014adam}
and depends on the pre-defined batch size, it's natural to come up with Minibatch K-Means.
With this procedure, the ``center-loss" further plays a role to mitigate the error caused different optimization
direction of K-Means and regular loss functions in each iteration of a whole epoch similar to \cite{hsu2018cnn}.

\subsection{Generalized KM-GAN}
Although common used datasets have obvious criterions to cluster, such as MNIST \cite{lecun1998gradient}
and CIFAR-10 \cite{netzer2011reading}. However, there exist datasets that do not have these
obvious criterions. For example, CelebA \cite{liu2015deep} and LFW \cite{learned2016labeled}
contains too many personalities and images for each personality are not enough
for generation tasks. It's even hard to find a suitable number for the pre-defined $k$.
In this case, operating K-Means to cluster features is too difficult.
To handle with such problem, we generalize regular KM-GAN with two regularization terms.
They act as constraints \cite{dou2017metric} on the whole class of real and fake images.
Before explaining the constraints, we define two necessary terms ${L}_{intra}$ and
${L}_{inter}$ used to generalize KM-GAN as follows:
\begin{align*}\label{regularization-terim}
  L_{intra}&=\sum_{{\mathbf{x}}_{i},{\mathbf{x}}_{j}\in{B}_{d}}\|D({\mathbf{x}}_{i})-D({\mathbf{x}}_{j}) \|_1 +
           \sum_{G({\mathbf{z}}_{i}),G({\mathbf{z}}_{j})\in{B}_{g}}\|D(G({\mathbf{z}}_{i}))-D(G({\mathbf{z}}_{j})) \|_1, \\
  \vspace{5pt}
  L_{inter}&=\sum_{{\mathbf{x}}_{i}\in{B}_{d}, G({\mathbf{z}}_{j})\in{B}_{g}}\|D({\mathbf{x}}_{i})-D(G({\mathbf{z}}_{j}))) \|_1,
\end{align*}
where ${B}_{d}$ and ${B}_{g}$ denote the corresponding batch of real samples
${\{{\mathbf{x}}_{i}\}}_{i=1}^{b}$ and generated samples ${\{G({\mathbf{z}}_{i})\}}_{i=1}^{b}$, respectively.

Then the objective functions of the discriminator and the generator become:
\begin{equation}\label{D-loss-G-loss-reg}
\begin{split}
  {L}_{D}  &= \min_{{\theta}_{D}}\|D(X) - C_{real} \|_2 - \|D(G(z)) - C_{gen} \|_2 + \lambda * ({L}_{intra}-{L}_{inter}), \\
  \vspace{5pt}
  {L}_{G}  &= \min_{{\theta}_{G}}\|D(G(z)) - C_{gen} \|_2 + \lambda * {L}_{inter}.
\end{split}
\end{equation}

In the case described above, the objective functions of regular KM-GAN are not effective enough since
they are dependent on $k$. However these two terms, one decreases intra-class distances of the whole
real and fake data in feature space while the other minimizes inter-class distance to approximate
data distribution, help to approximate the data distribution as a whole class.
With above regularization terms, experimental results also show that the final centers reduce to
the same one whatever the pre-defined $k$ is (such as $k=10$ or $k=20$), which coincides
with the goal of these regularization terms. This implies that KM-GAN could adapt to more general
scenarios with them. We use the hyperparameter $\lambda$ in experiments to balance the
regular loss functions and these two regularization terms.

\section{Experiments}
In this section, we first conduct experiments on a synthetic dataset to show
the capability of the discriminator of KM-GAN to represent features.
Then we qualitatively and quantitatively show that KM-GAN is able to generate realistic
and diverse images on real-world datasets including MNIST, Fashion-10, CIFAR-10 and CelebA.
Details about these datasets are shown in Table \ref{details-of-datasets}. Note that the
hyperparameter $\lambda$ is set to 0 in experiments except CelebA, where it is set to 5.

\begin{table}[!htbp]
\centering
\tabcolsep 8pt
\label{details-of-datasets}
\scalebox{1.0}[1.0]{
\begin{tabular}{|l|c|c|c|}
  \hline
  Dataset   & Numbers of Images & Feature Dimensions & Classes   \\
  \hline
  Synthesis & 10,000  & 100                       & 4   \\ \hline
  MNIST     & 70,000  & 28$\times$28              & 10  \\ \hline
  Fashion-10& 70,000  & 28$\times$28              & 10  \\ \hline
  CIFAR-10  & 70,000  & 32$\times$32$\times$3     & 10  \\ \hline
  CelebA    & 202,599 & 64$\times$64$\times$3     & No  \\
  \hline
\end{tabular}
}
\caption{Details of synthetic data and real-world datasets.}
\end{table}

\subsection{KM-GAN on Synthetic Data}
\begin{figure*}
    \centering
    \begin{subfigure}[t]{0.19\textwidth}
        \includegraphics[width=\textwidth]{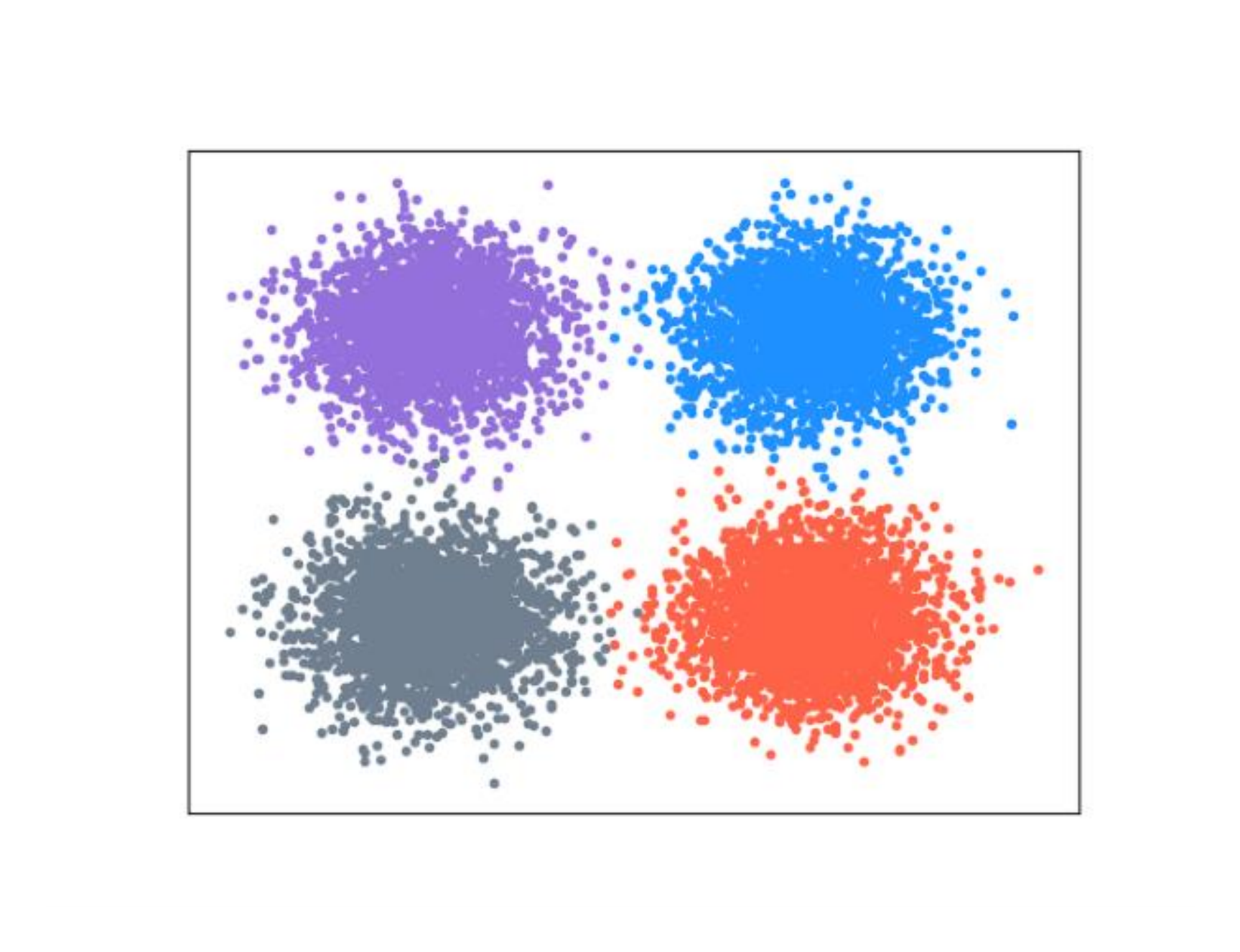}
        \caption{Real data}
        \label{synthetic-data}
    \end{subfigure}
    \begin{subfigure}[t]{0.19\textwidth}
        \includegraphics[width=\textwidth]{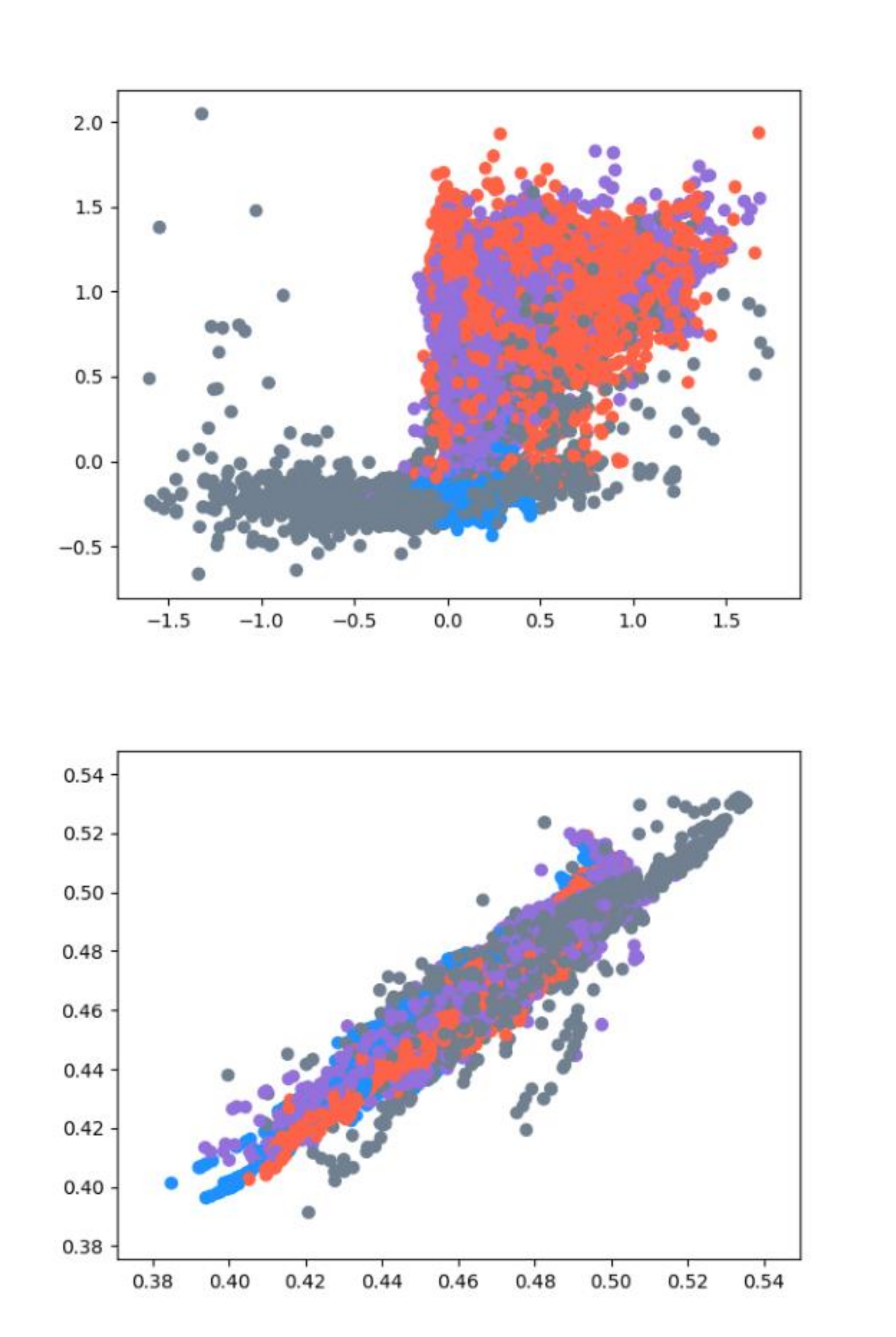}
        \caption{Epoch 0}
        \label{epoch-0}
    \end{subfigure}
    \begin{subfigure}[t]{0.19\textwidth}
        \includegraphics[width=\textwidth]{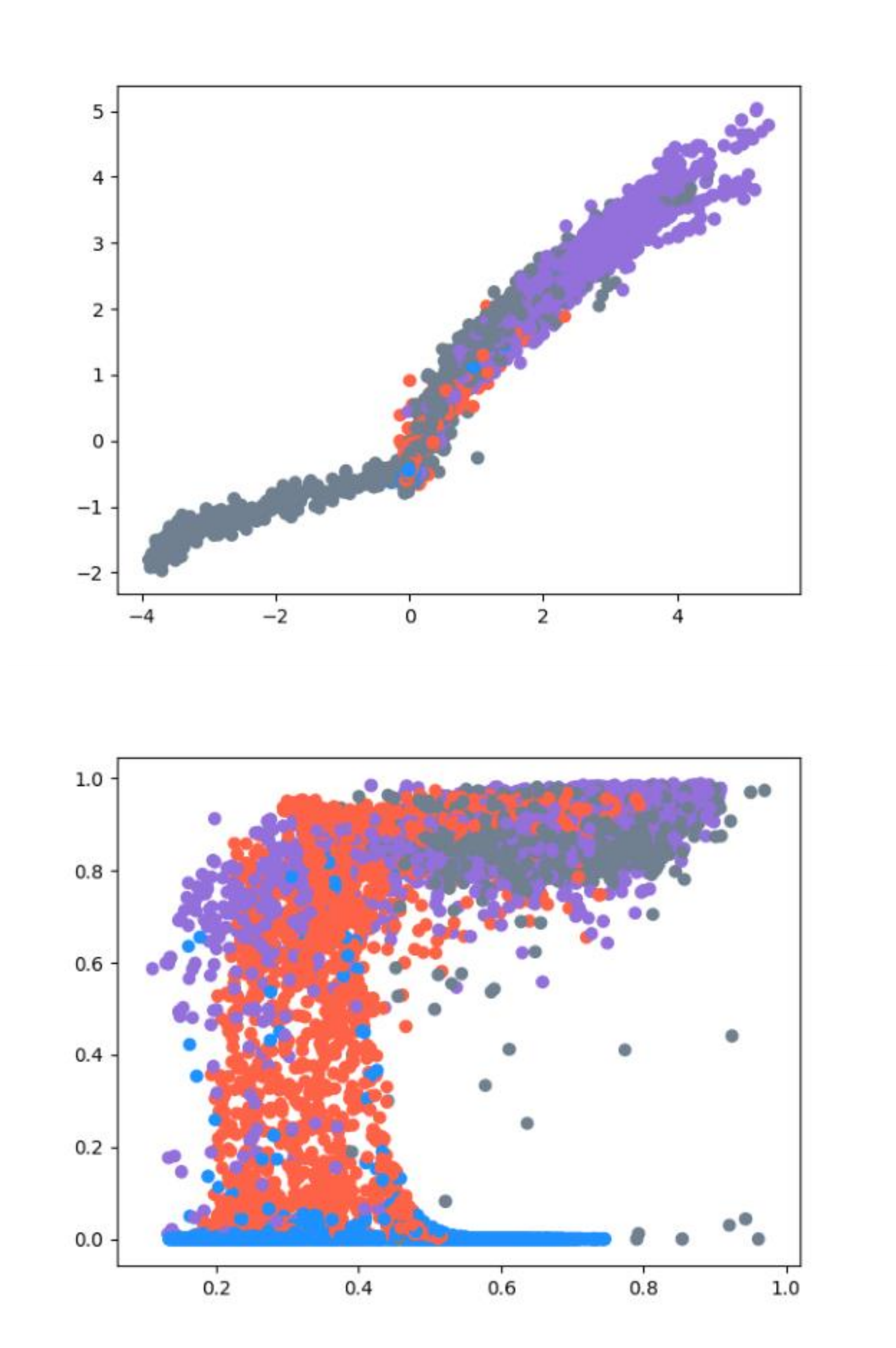}
        \caption{Epoch 50}
        \label{epoch-50}
    \end{subfigure}
    \begin{subfigure}[t]{0.19\textwidth}
        \includegraphics[width=\textwidth]{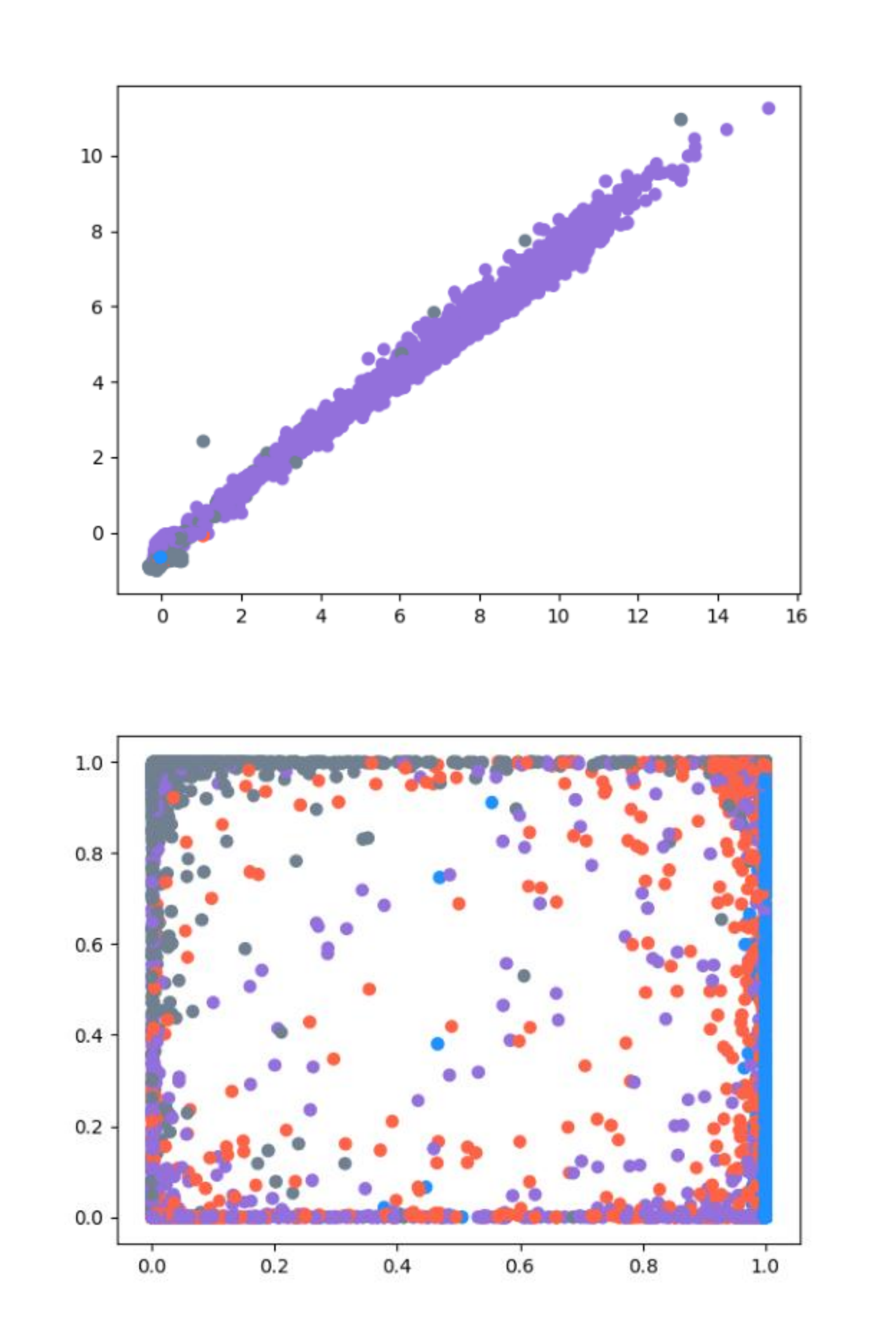}
        \caption{Epoch 150}
        \label{epoch-150}
    \end{subfigure}
    \begin{subfigure}[t]{0.19\textwidth}
        \includegraphics[width=\textwidth]{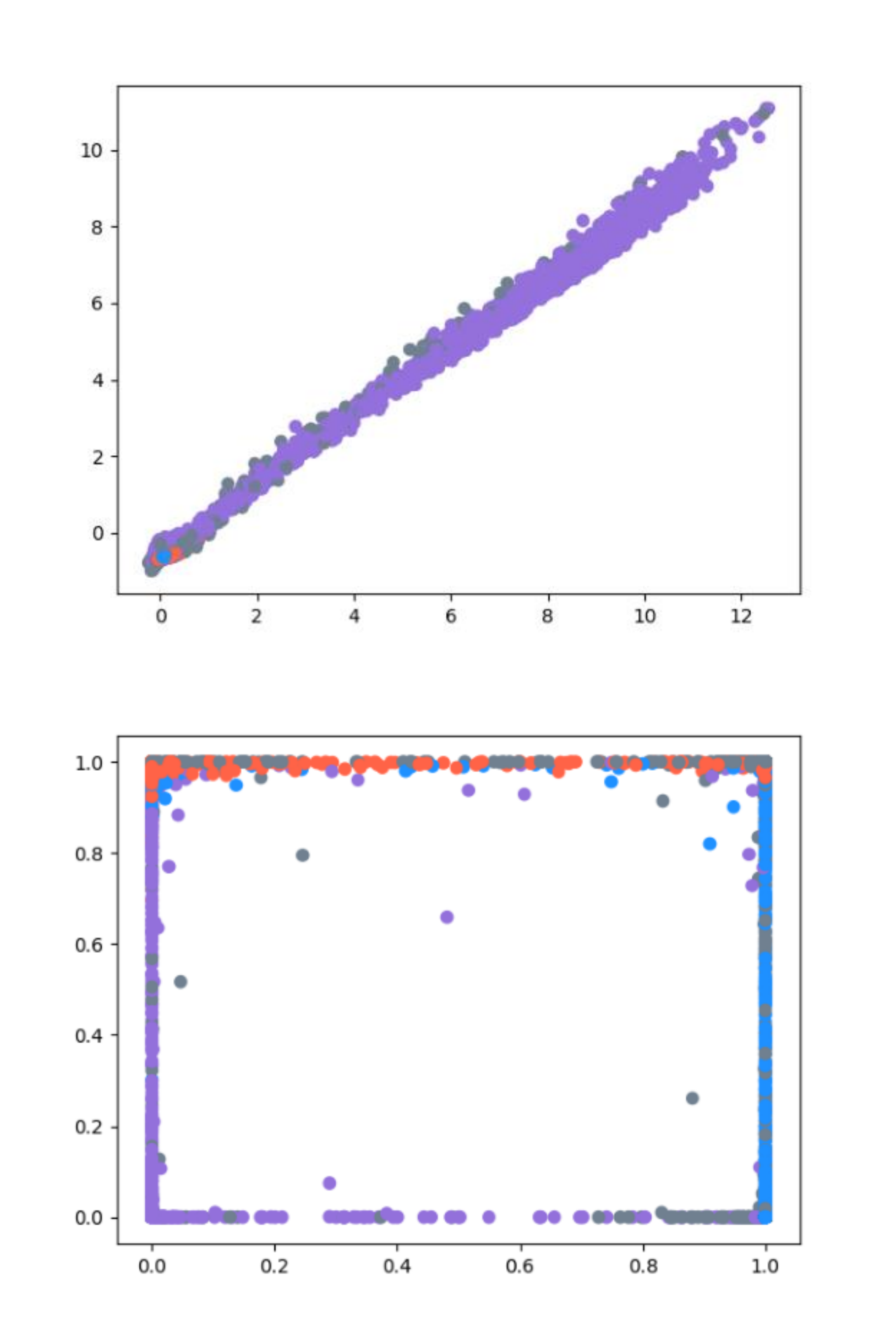}
        \caption{Epoch 200}
        \label{epoch-200}
    \end{subfigure}

    \caption{Subfigure (a) is the visualization of intrinsic 2-dimension structure of synthetic data.
Subfigures (b)-(e) are visualization of features on the top layer of corresponding discriminators of DCGAN and KM-GAN in the
training process on synthetic dataset. The four kinds of colored points represent different categories. Obviously,
the features of KM-GAN could separate most of them while that of DCGAN is ineffective.}
\label{feature_of_discriminator_comparison}
\end{figure*}

As we can see from loss functions of KM-GAN, features of the discriminator play an important
role not only on the objective function of the discriminator itself, but also on that of the generator.
To demonstrate that features of KM-GAN are really representative, we compare with that of DCGAN which
shows its capability to do representation learning by conducting classification experiments using its trained discriminator.
The obtained features of KM-GAN and DCGAN on a synthetic dataset in the training process are shown in Fig. \ref{feature_of_discriminator_comparison}.

\makeatletter\def\@captype{figure}\makeatother

The synthetic dataset consists of $10,000$ points that belong to $\mathbb{R}^{100}$
and has ``K-Means-friendly" \cite{yang2017towards} structure in a two-dimensional domain which
we could not observe. In fact, we first choose four two-dimensional gaussian distributions with
different means and covariance matrices as in Fig. \ref{synthetic-data}.
Then sample $2,500$ points from each distribution and map them into $\mathbb{R}^{100}$ through a
mapping function $\mathcal{M}$, which is realized by a non-linear neural network showed in
Table \ref{mapping-function}. In this experiment, we set ${d}_{round}$ as $0$, and the network
structures of DCGAN\cite{goodfellow2014generative} and KM-GAN are the same and both shown in Table \ref{mapping-function}.
The visualization of the features learned in the training process by discriminators of these two models are showed in Fig. \ref{feature_of_discriminator_comparison}. Compared with features of DCGAN, those of our proposed KM-GAN
are obviously more representative to show the intrinsic structure although they are both capable to generate
high-quality images on real-world datasets.

\makeatletter\def\@captype{table}\makeatother
\begin{center}
\label{mapping-function}
\begin{tabular}{lll}
  \hline
  Mapping function $\mathcal{M}$ &  Generator      &Discriminator                             \\
  \hline
  \hline
  Input ${\{{\mathbf{h}}_{i}\}}_{i=1}^{n} \in {\mathbb{R}}^{2}$  &  Input ${\{{\mathbf{z}}_{i}\}}_{i=1}^{n} \in {\mathbb{R}}^{100}$     &Input ${\{{\mathbf{x}}_{i}\}}_{i=1}^{n} \in {\mathbb{R}}^{100}$                                                        \\
  FC 10 Sigmoid    &                     FC 10  ReLU BN  & FC 100 ReLU BN                                                 \\
  FC 100 Sigmoid   &                     FC 50  ReLU BN  & FC 50 ReLU BN                                                  \\
   ${\{{\mathbf{x}}_{i}\}}_{i=1}^{n} \in {\mathbb{R}}^{100}$     & FC 100  ReLU BN      &     FC 10 ReLU BN         \\
     &    ${\{G({\mathbf{z}}_{i})\}}_{i=1}^{n} \in {\mathbb{R}}^{100}$   &  FC 2 Sigmoid \\
       &          & ${\{D({\mathbf{x}}_{i}), D(G({\mathbf{z}}_{i}))\}}_{i=1}^{n} \in {\mathbb{R}}^{2}$           \\
  \hline
\end{tabular}
\caption{Architectures of $\mathcal{M}$, generator and discriminator.}
\end{center}

\subsection{KM-GAN on MNIST}
MNIST \cite{lecun1998gradient} dataset has $70,000$ gray images of handwritten digits of size
$28\times28$. We first conduct experiments to compare KM-GAN with its reduced version
which operates K-Means in pixel space as introduced in Algorithm \ref{alg-reduce}.
Then we improve KM-GAN with weight-clipping which stabilizes the training process.
The network structures of KM-GAN for training MNIST are the same as that of DCGAN and
hyperparameter ${d}_{round}$ is set as $10,000$.

\subsubsection{Feature Space vs. Original Space}
\begin{figure*}
    \centering
    \begin{subfigure}[t]{0.32\textwidth}
        \includegraphics[width=\textwidth]{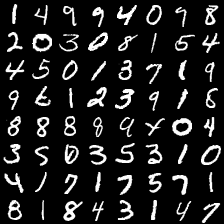}
        \caption{Real data}
        \label{mnist}
    \end{subfigure}
    \begin{subfigure}[t]{0.32\textwidth}
        \includegraphics[width=\textwidth]{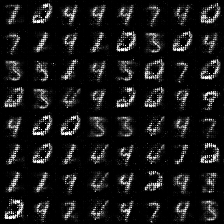}
        \caption{Reduced KM-GAN}
        \label{mnist-kmgan-original}
    \end{subfigure}
    \begin{subfigure}[t]{0.32\textwidth}
        \includegraphics[width=\textwidth]{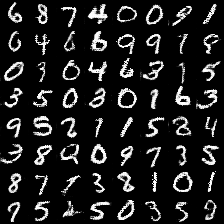}
        \caption{KM-GAN}
        \label{mnist-kmgan}
    \end{subfigure}

    \caption{Comparison of generated samples of KM-GAN and reduced version of KM-GAN on MNIST dataset.}
\label{mnist-original-feture-space}
\end{figure*}

To demonstrate the effect of carrying out K-Means in feature space rather than pixel space,
we compare KM-GAN with reduced KM-GAN, in which we operate K-Means in pixel space and cluster original data.
Indeed, computations in K-Means appear to increase quickly as the dimensionality of data increases
when experimenting with reduced KM-GAN. However, the capability of dimensionality reduction of KM-GAN avoids
such computational difficulties. In the following, we further qualitatively show the advantage of operating
K-Means on latent space as exhibited images in Fig. \ref{mnist-original-feture-space}.

\begin{algorithm}
\caption{Training algorithm for reduced KM-GAN}
\label{alg-reduce}
\begin{algorithmic}
\REQUIRE Real Images $\mathbf{X}$, noise distribution${P}_{\mathbf{Z}}$,
pre-defined number of classes $k$, number of iterations $T$, batch size $b$
and hyper-parameters of Adam $\alpha, {\beta}_{1}, {\beta}_{2}$
\ENSURE Generated samples $G(\mathbf{z})$

\STATE Initialize $k$ real centers $\widetilde{C}$ of data $\mathbf{X}$ by K-Means++
\STATE Initialize parameters of $D$ and $G$ networks
\FOR{$t=1:T$}
\STATE Sample a batch $\{{\mathbf{x}}_{i}\}_{i=1}^{b}$ from real data $\mathbf{X}$
\STATE Sample a batch $\{{\mathbf{z}}_{i}\}_{i=1}^{b}$ from noise distribution ${P}_{\mathbf{Z}}$
\STATE Obtain features $\{D({\mathbf{x}}_{i})\}_{i=1}^{b}$ and $\{D(G({\mathbf{z}}_{i}))\}_{i=1}^{b}$
\STATE Obtain clustering labels according to Euclidean distance with current $k$ centers
\vspace{2pt}
\STATE ${\widetilde{L}}_{center} = \|D({\widetilde{C}}_{real})-D({\widetilde{C}}_{gen})\|_1$
\STATE ${grad}_{{\theta}_{d}, {\theta}_{g}}= {\triangledown}_{{\theta}_{d},
{\theta}_{g}}{\widetilde{L}}_{center}$
\vspace{2pt}
\STATE ${\theta}_{d}, {\theta}_{g}=$Adam(${grad}_{{\theta}_{d}, {\theta}_{g}}, {\theta}_{d}, {\theta}_{g}, \alpha, {\beta}_{1}, {\beta}_{2}$)
\vspace{2pt}
\STATE ${\widetilde{L}}_{D} = \|D(\mathbf{x})-D({\widetilde{C}}_{real})\|_2$
\STATE ${grad}_{{\theta}_{d}}= {\triangledown}_{{\theta}_{d}}{\widetilde{L}}_{D}$
\vspace{2pt}
\STATE ${\theta}_{d}=$Adam(${grad}_{{\theta}_{d}}, {\theta}_{d}, \alpha, {\beta}_{1}, {\beta}_{2}$)
\vspace{2pt}
\STATE $ {\widetilde{L}}_{G} = \|D(G(\mathbf{z}))-D({\widetilde{C}}_{gen})\|_2$
\STATE ${grad}_{{\theta}_{g}}= {\triangledown}_{{\theta}_{g}}{\widetilde{L}}_{G}$
\vspace{2pt}
\STATE ${\theta}_{g}=$Adam(${grad}_{{\theta}_{g}}, {\theta}_{g}, \alpha, {\beta}_{1}, {\beta}_{2}$)
\vspace{2pt}
\STATE Update centers in original pixel space via K-Means objective in Equation \ref{objective of kmeans}
\ENDFOR
\end{algorithmic}
\end{algorithm}

From images in Fig. \ref{mnist-kmgan-original} and Fig. \ref{mnist-kmgan}, it is obvious
that the quality of generated digits is significantly better when clustering is operated
in the feature space. Regular KM-GAN successfully generates realistic handwritten digits not only in
different classes and angles while reduced KM-GAN even suffers mode collapse, i.e.,
most of generated images are similar or identical.

\makeatletter\def\@captype{figure}\makeatother

\subsubsection{Improvement on KM-GAN}
Although KM-GAN is proven to be capable to generate realistic and diverse images, it still fails to generate
images sometimes. So we utilize a common technique called weight clipping to constrain parameters of the
discriminator (feature extractor). Specifically, we clamp the weights of $D$ to a fixed box so that it could
only output values in a certain range. The technique further guarantees the property that points close in pixel
space are not far away from each other after mapped into feature space.

\begin{figure*}
    \centering
    \begin{subfigure}[t]{0.32\textwidth}
        \includegraphics[width=\textwidth]{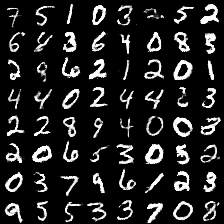}
        \caption{DCGAN}
        \label{mnist-dcgan}
    \end{subfigure}
    \begin{subfigure}[t]{0.32\textwidth}
        \includegraphics[width=\textwidth]{mnist-kmgan.jpg}
        \caption{KM-GAN}
        \label{mnist-kmgan-1}
    \end{subfigure}
    \begin{subfigure}[t]{0.32\textwidth}
        \includegraphics[width=\textwidth]{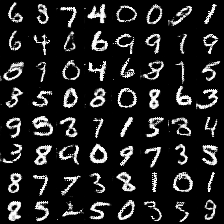}
        \caption{KM-GAN with weight clipping}
        \label{mnist-kmgan-wl}
    \end{subfigure}

    \caption{Subfigures (a) and (b) compare generated images of DCGAN and KM-GAN on MNIST dataset, and subfigure (c)
    shows generated images of KM-GAN improved with weight clipping.}
\label{mnist-weight-clipping}
\end{figure*}

As synthesized images shown in Fig. \ref{mnist-dcgan} and Fig. \ref{mnist-kmgan-1},
the performance of KM-GAN without weight clipping is already competitive with DCGAN on MNIST dataset.
This demonstrates that the utilization of clustering labels successfully replaces the role of real
labels to direct generating process and encourages us to pay more attention to un-conditional generative models.
What's more, to stabilize the three-step alternating optimization process, we equip KM-GAN with weight clipping and
the clipping threshold is set to $[-1, 1]$. The synthesized images shown in Fig. \ref{mnist-kmgan-wl} are
competitive or even better than KM-GAN without weight clipping.

\subsection{KM-GAN on Fashion-10}
\begin{figure*}[!h]
    \centering
    \begin{subfigure}[t]{0.27\textwidth}
        \includegraphics[width=\textwidth]{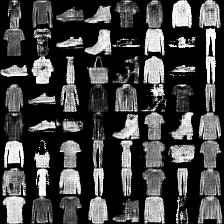}
        \caption{DCGAN}
        \label{fashion-dcgan}
    \end{subfigure}
    \begin{subfigure}[t]{0.27\textwidth}
        \includegraphics[width=\textwidth]{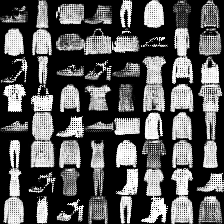}
        \caption{KM-GAN}
        \label{fashion-kmgan}
    \end{subfigure}
    \begin{subfigure}[t]{0.43\textwidth}
        \includegraphics[width=\textwidth]{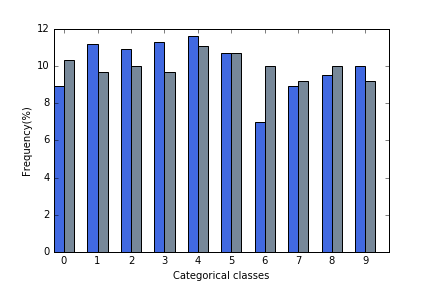}
        \caption{Frequency chart}
        \label{fre-pic}
    \end{subfigure}

    \caption{Evaluation of synthesized images of KM-GAN on Fashion-10 dataset. Subfigures
    (a) and (b) exhibit a random batch of generated images of DCGAN and KM-GAN, and subfigure
    (c) shows the distributions of generated images of DCGAN and KM-GAN, respectively.}
\label{fashion}
\end{figure*}

Fashion-10 dataset, consisting of various types of more complicated fashion products rather
than handwritten digits, has the same number of images as MNIST and the size of each image
is also $28\times28$. So we use the same architecture as used on MNIST to examine KM-GAN on Fashion-10.
The hyperparameter ${d}_{round}$ is also the same as on MNIST. From the experimental results shown in
Fig. \ref{fashion-dcgan} and Fig. \ref{fashion-kmgan}, the quality of synthesized images of KM-GAN is comparable to that of DCGAN.

To further quantitatively show that our proposed method is also capable to generate diverse
images without the help of one-hot real labels, we train a three-layer convolutional classifier
on Fashion-10 separately ($97\%$ accuracy on training set and $91\%$ on test set)
and use the classifier to classify $5,000$ synthesized images of KM-GAN and DCGAN.
The result of the frequency of each class is shown in Fig. \ref{fre-pic}.
Indeed, since Fashion-10 equally contains images of each class, conditional models
easily generate images equally for each class with the help of real labels.
So we compare with results of DCGAN to further show that KM-GAN is also capable achieve this.
Specifically, in the frequency chart of generated images, numbers $0\sim9$ denote $10$ classes of
the dataset and two colors, ``blue" and ``gray", represent results of KM-GAN and
DCGAN, respectively. From the class distributions, most classes are generated with probability
close to $10\%$ by KM-GAN except the class ``shirts", which is under-represented with $7.0\%$.
We infer this is because that ``shirts" are very similar to ``T-shirts" and ``pullovers".

\subsection{KM-GAN on CIFAR-10}

CIFAR-10 \cite{netzer2011reading} is a dataset with $60,000$ RGB images of size $32\times 32$
in 10 classes. There are $6,000$ images in each class with $5,000$ for training and $1,000$ for testing.
All these images are used here to train KM-GAN. The network structures are shown in
Table \ref{architecture-cifar10} and we set ${d}_{round}$ to $20,000$ and clipping threshold to $[-0.01, 0.01]$.

\begin{figure*}[!h]
    \centering
    \begin{subfigure}{0.325\textwidth}
        \includegraphics[width=\textwidth]{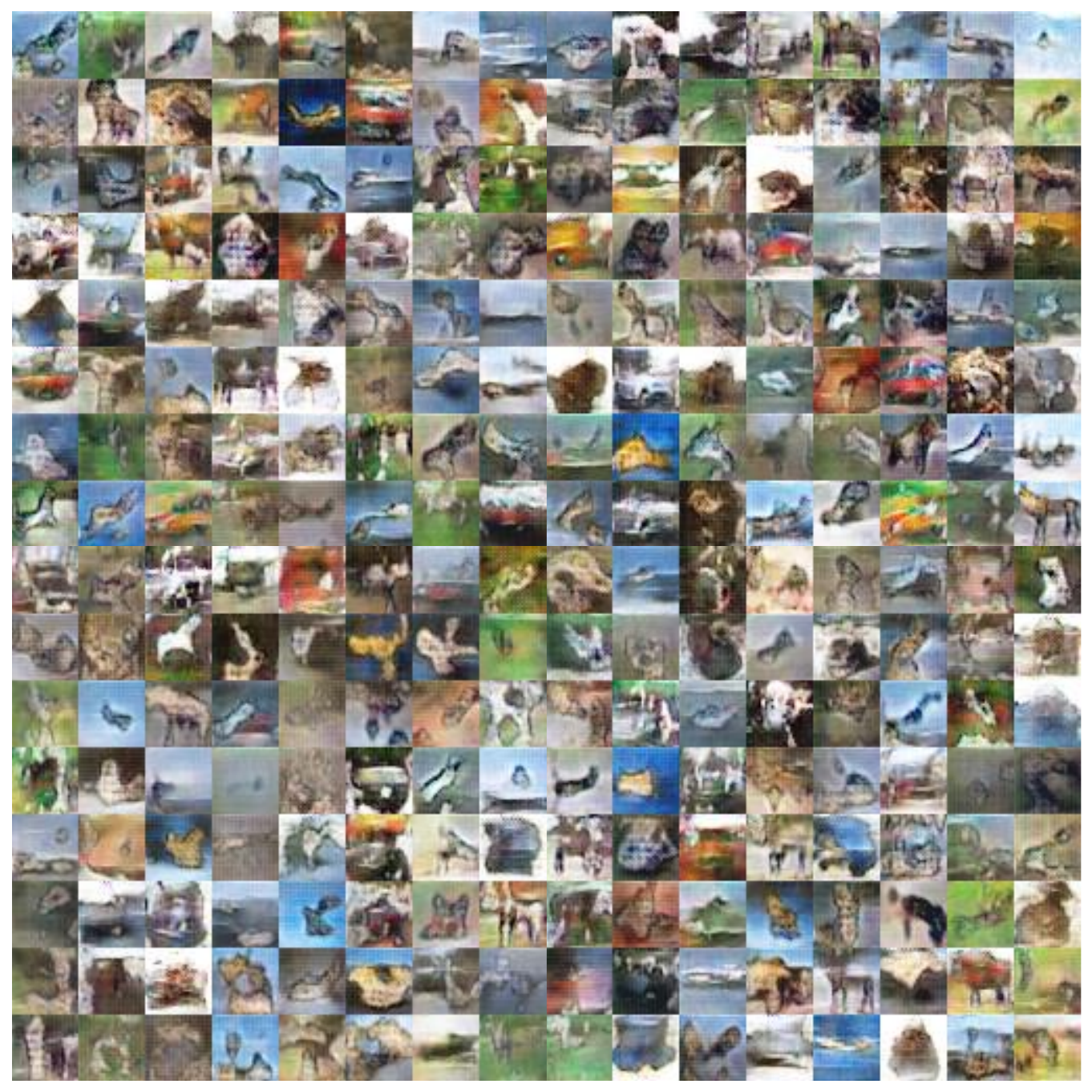}
        \caption{MBGAN}
        \label{cifar-mbgan}
    \end{subfigure}
    \begin{subfigure}{0.325\textwidth}
        \includegraphics[width=\textwidth]{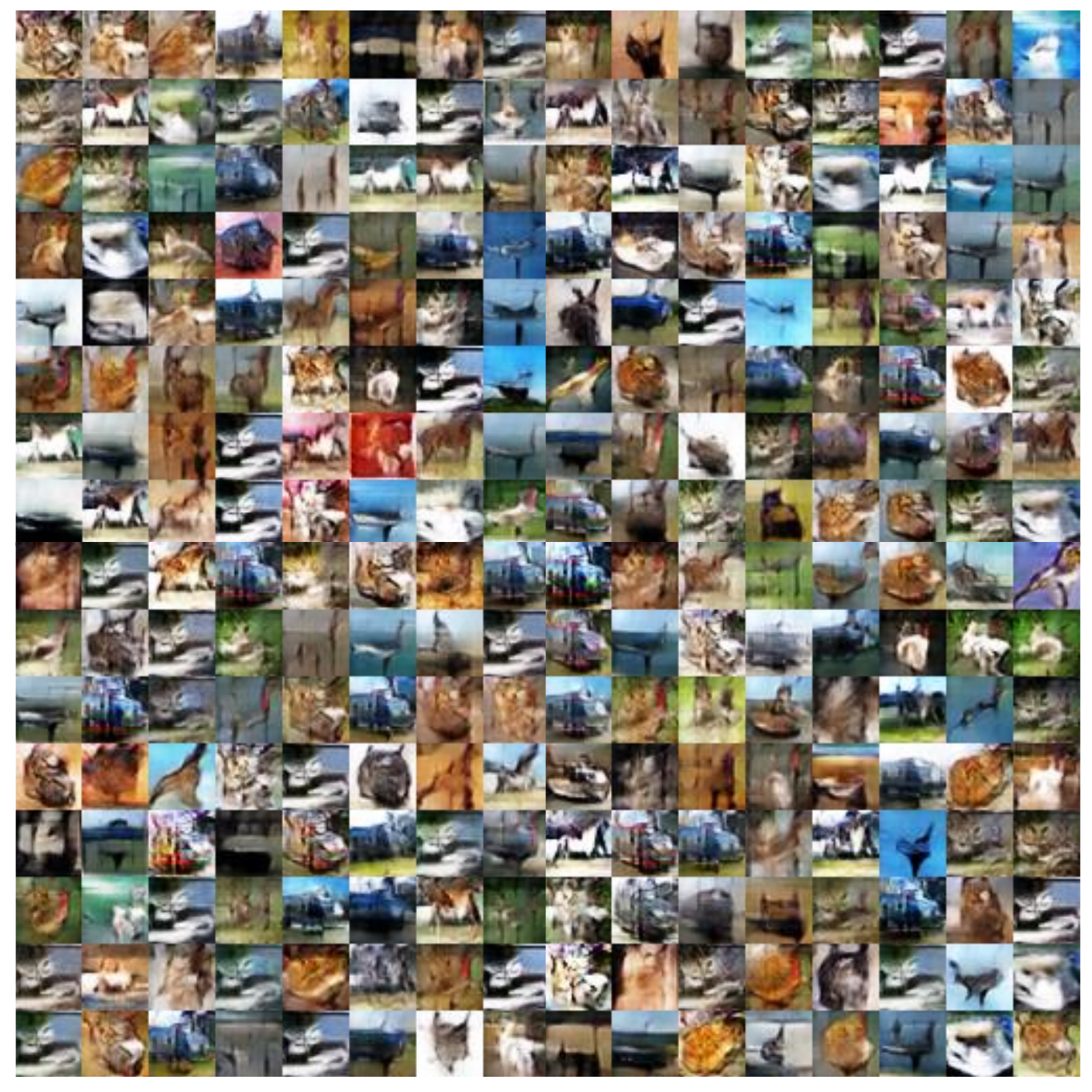}
        \caption{DCGAN}
        \label{cifar-dcgan}
    \end{subfigure}
    \begin{subfigure}{0.325\textwidth}
        \includegraphics[width=\textwidth]{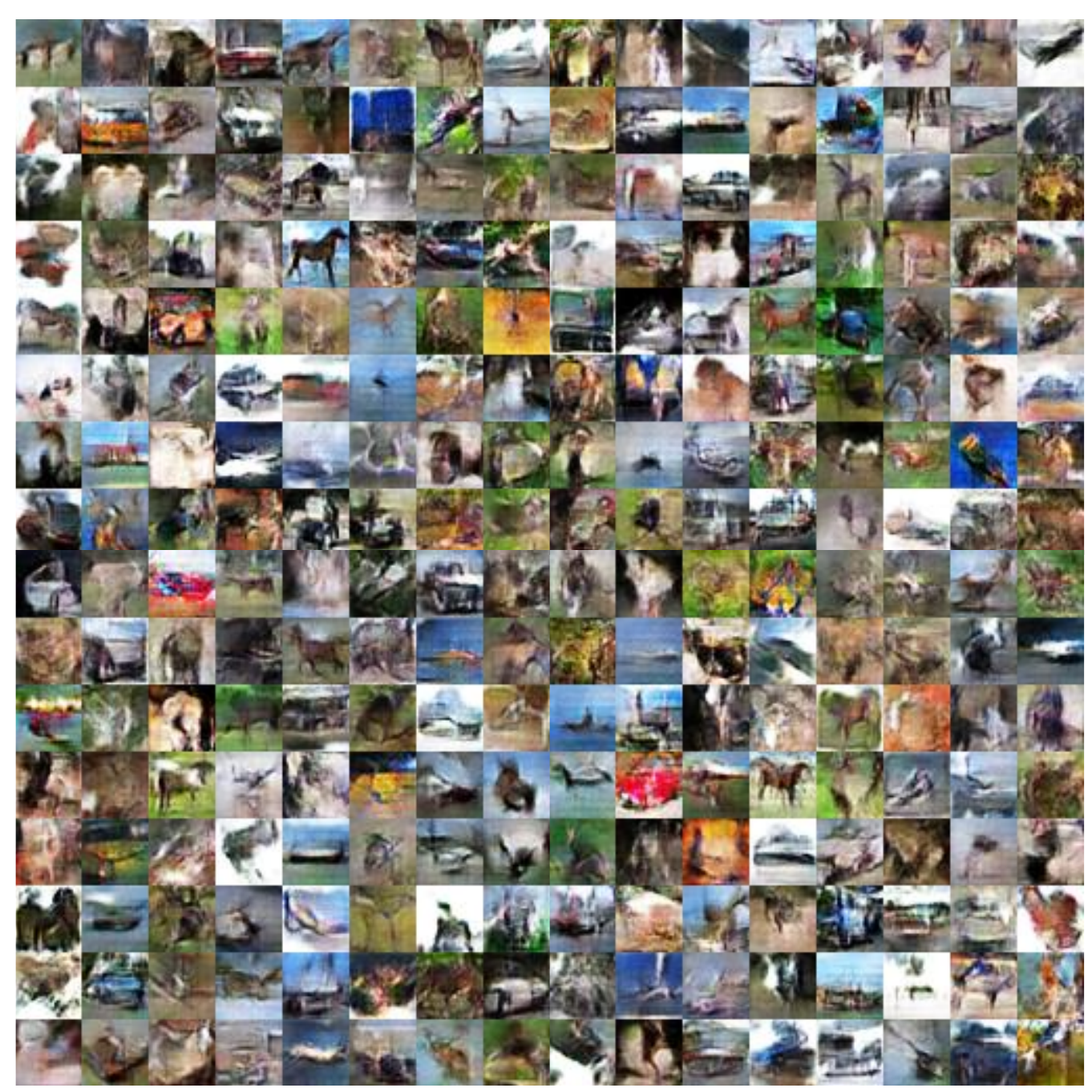}
        \caption{KM-GAN}
        \label{cifar-kmgan}
    \end{subfigure}

    \caption{Comparison of generated samples of MBGAN, DCGAN and KM-GAN on CIFAR-10.}
\label{cifar}
\end{figure*}

We first evaluate the generated images of KM-GAN on CIFAR-10 dataset and show the experimental results
in Fig. \ref{cifar}. To demonstrate the capability of our proposed objective functions, we compare with
MBGAN which also proposes different objective functions from the perspective of DML. Results shows that synthesized
images of KM-GAN are obviously more realistic and meaningful. We further compare with DCGAN and there is
no visual difference between the quality of synthesized images of these two models, which again demonstrates
the effectiveness of KM-GAN.

\makeatletter\def\@captype{table}\makeatother
\begin{center}
\caption{Architectures of generator and discriminator on CIFAR-10.}
\label{architecture-cifar10}
\begin{tabular}{ll}
  \hline
  Generator      &Discriminator                                    \\
  \hline
  \hline
  Input ${\{{\mathbf{z}}_{i}\}}_{i=1}^{n} \in {\mathbb{R}}^{100}$  &
  Input ${\{{\mathbf{x}}_{i}\}}_{i=1}^{n} \in {\mathbb{R}}^{64\times 64 \times 3}$   \\
  \hline
  FC $4\times4\times512$ BN ReLU         & $5\times5$ Conv                           \\
                                         & 64 stride 2 ReLU                          \\
  \hline
  $5\times5$ Upconv                      & $5\times5$ Conv                           \\
  256 stride 2 BN ReLU                   & 128 stride 2 BN ReLU                      \\
  \hline
  $5\times5$ Upconv                      & $5\times5$ Conv                           \\
  128 stride 2 BN ReLU                   & 256 stride 2 BN ReLU                      \\
  \hline
  $5\times5$ Upconv                      & $5\times5$ Conv                           \\
  64 stride 2 BN ReLU                    & 512 stride 2 BN ReLU                      \\
  \hline
  $5\times5$ Upconv                      & FC 4096 BN ReLU                           \\
  3 Stride 2 Sigmoid                     &                                           \\
  \hline
                                         & FC 100 BN ReLU                            \\
  \hline
\end{tabular}
\end{center}

Since we use clustering labels of K-Means to replace one-hot real labels in KM-GAN, i.e., a purely un-supervised
training, we quantitatively evaluate the diversity of images synthesized by our model
with another index called inception score \cite{salimans2016improved} on CIFAR-10 dataset.
The index applies Inception model \cite{szegedy2016rethinking} to every generated image and
computes the following metric:
\begin{equation}
\mathbf{IS}(G(\mathbf{z})) = \exp({\mathbb{E}}_{\mathbf{z}}\mathbf{KL}(p(y|G(\mathbf{z}))\|p(y))).
\label{inception-score-equation}
\end{equation}

Indeed, the main idea of Equation \ref{inception-score-equation} is that diverse generated images which contain meaningful objects
are supposed to have a conditional label distribution $p(y|G(\mathbf{z}))$ with low entropy and a
marginal distribution $\int p(y|G(\mathbf{z}))d\mathbf{z}$ with high entropy. As in Table \ref{inception-score},
we report inception scores of both conditional and un-conditional models to characterize the performance of KM-GAN.
Specifically, WGAN, Improved GANs, and MIX+WGAN are trained without feeding real labels, while ALI is itself
an un-conditional model utilizing an auto-encoder to assist the generator to approximate target distribution.
Obviously, KM-GAN performs much better than these models which demonstrates the effectiveness of KM-GAN.
We then compare with two conditional methods based on DML, MLGAN and MBGAN. KM-GAN also works better than
than them. Furthermore, we compare with DCGAN, a very stable and common used conditional method in the research
field of GANs. Results show that KM-GAN are comparable to DCGAN just like from above synthesized images.
We infer that this is because synthesized images of KM-GAN shown in Fig. \ref{cifar-kmgan} are more meaningful while
the backgrounds of generated images of DCGAN shown in Fig. \ref{cifar-dcgan} are more clear.
\begin{center}
\caption{Inception Score on CIFAR-10 Dataset.}
\label{inception-score}
\begin{tabular}{llr}
  \hline
  &  Model                 & Inception Score\\
  \hline
  \hline
  &  MBGAN                                                                                 & 4.27 $\pm$ 0.07                      \\
  Conditional  &  MLGAN-clipping \cite{dou2017metric}                                      & 5.23 $\pm$ 0.29                      \\
  Models &  DCGAN                                                                          & 5.37 $\pm$ 0.06                      \\
  \hline
  \hline
  &  MIX + WGAN \cite{arora2017generalization}                                             & 4.04 $\pm$ 0.07                      \\
  Un-conditional  &  Improved GANs \cite{salimans2016improved}                             & 4.36 $\pm$ 0.04                      \\
  Models &  ALI \cite{dumoulin2016adversarially} (from \cite{pu2017adversarial})           & 4.79                                 \\
  &  Wasserstein GANs \cite{arjovsky2017wasserstein} (from \cite{arora2017generalization}) & 3.82 $\pm$ 0.06                      \\
  &  $\mathbf{KM-GAN}$                                                                     & 5.39 $\pm$ 0.05                      \\
  \hline
\end{tabular}
\end{center}

\subsection{KM-GAN on CelebA}
CelebA \cite{liu2015deep}, as a large-scale face dataset, contains more than $200,000$ RGB face
images from $10,177$ celebrity identities, and there are 40 binary attributes
and 5 landmarks for each image. In this experiment, we crop images into $64\times 64$,
and the network structures are shown in Table \ref{architecture-celeba}. The hyperparameters ${d}_{round}$ and
clipping threshold are set the same as in CIFAR-10 dataset. Besides, we set $\lambda$ set as 5 since
we could not find an appropriate $k$ for this CelebA while other datasets have determinate categories.
Following are samples generated by DCGAN and KM-GAN, respectively.

\makeatletter\def\@captype{table}\makeatother
\begin{center}
\caption{Architectures of generator and discriminator on CelebA.}
\label{architecture-celeba}
\begin{tabular}{ll}
  \hline
  Generator      &Discriminator                                   \\
  \hline
  \hline
  Input ${\{{\mathbf{z}}_{i}\}}_{i=1}^{n} \in {\mathbb{R}}^{100}$  &
  Input ${\{{\mathbf{x}}_{i}\}}_{i=1}^{n} \in {\mathbb{R}}^{64\times 64 \times 3}$   \\
  \hline
  FC $4\times4\times512$ BN ReLU         & $5\times5$ Conv                           \\
                                         & 64 stride 2 BN ReLU                       \\
  \hline
  $5\times5$ Upconv                      & $5\times5$ Conv                           \\
  256 stride 2 BN ReLU                   & 128 stride 2 BN ReLU                      \\
  \hline
  $5\times5$ Upconv                      & $5\times5$ Conv                           \\
  128 stride 2 BN ReLU                   & 256 stride 2 BN ReLU                      \\
  \hline
  $5\times5$ Upconv                      & $5\times5$ Conv                           \\
  128 stride 2 BN ReLU                   & 512 stride 2 BN ReLU                      \\
  \hline
  $5\times5$ Upconv                      & FC 100 Sigmoid                            \\
  3   Stride 2 Tanh                      &                                           \\
  \hline
\end{tabular}

\end{center}
\begin{figure*}[!htbp]
    \centering
    \begin{subfigure}{0.48\textwidth}
        \includegraphics[width=\textwidth]{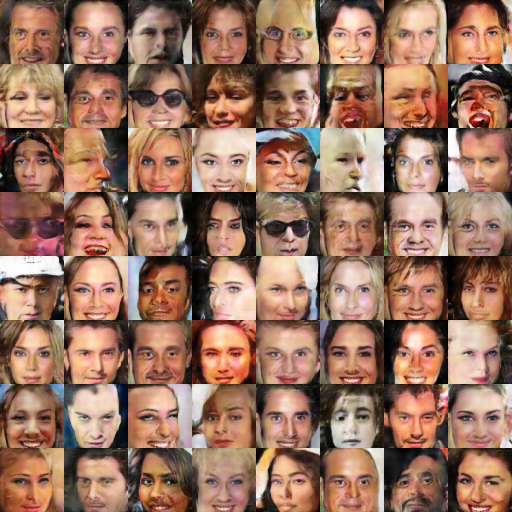}
        \caption{DCGAN}
        \label{celeba-dcgan}
    \end{subfigure}
    \begin{subfigure}{0.48\textwidth}
        \includegraphics[width=\textwidth]{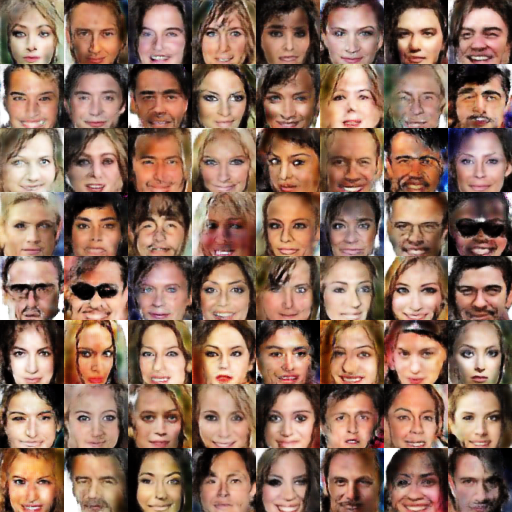}
        \caption{KM-GAN}
        \label{celeba-kmgan}
    \end{subfigure}

    \caption{Comparison of generated samples of DCGAN and KM-GAN on CelebA.}
\label{celeba}
\end{figure*}

From results in Fig. \ref{celeba}, we see that
KM-GAN also works well on CelebA.
Then we interpolate synthesized images to demonstrate the generalization capability of KM-GAN
rather than only generating the training face images. We first interpolate
$\mathbf{z}\in {\mathbb{R}}^{100}$ and then map interpolated $\mathbf{z}$ with the generator.
The results are as shown in Fig. \ref{celeba-interpolation}.
The leftmost and rightmost images are mapped from ${\mathbf{z}}_{0}$ and
${\mathbf{z}}_{1} $, respectively. The other images are generated from
${\mathbf{z}}_{\beta}=\beta {\mathbf{z}}_{0} + (1-\beta){\mathbf{z}}_{1}$
$(\beta \in [0,1])$, i.e., interpolations of corresponding noise vectors.
As examples in Fig. \ref{celeba-interpolation}, generated images change smoothly
from leftmost to rightmost. Indeed, we choose features of faces, including
hair color, angles of faces, with or without eyeglasses and some other special features,
to exhibit the continuous change clearly.
Especially, on the first row, the face of a smiling woman with golden hair
transitions to the face of a seriously man with dark hair slowly.
In addition, on the second row, the face of a woman with dark hair and
close mouth changes to the face of a smiling woman with golden hair.
These interpolations indicate that our proposed KM-GAN is able to generate images
continuously instead of only memorizing training data.

\makeatletter\def\@captype{figure}\makeatother
\begin{center}
\includegraphics[height=4.9cm]{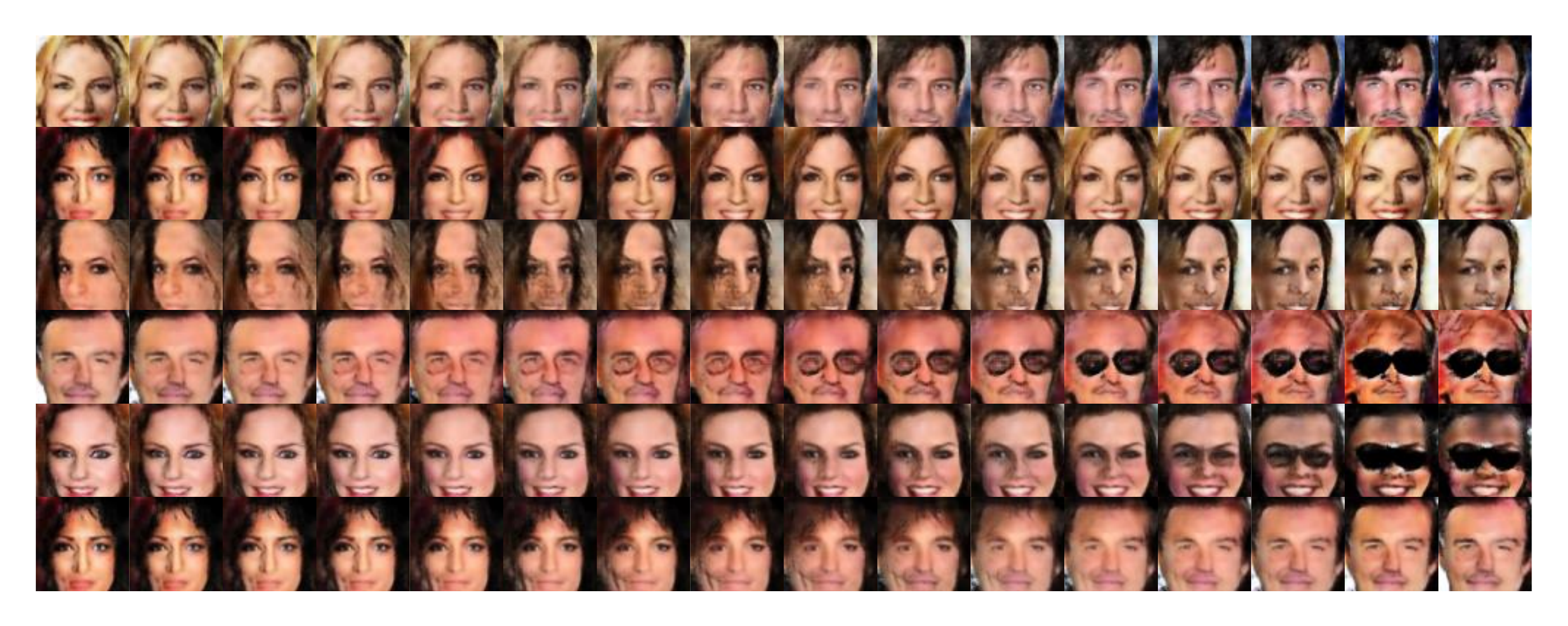}
\captionsetup{font={small}}
\caption{Interpolations of generated images on CelebA dataset.}
\label{celeba-interpolation}
\end{center}

\section{Conclusion}
In this paper, we propose an un-conditional extension of GANs, called KM-GAN, by fusing
with the idea of K-Means and utilizing the clustering results to propose
objective functions that direct the generating process. The purpose is
to replace the role of one-hot real labels with the clustering results, which generalizes
GANs to applications where real labels are expensive or impossible to obtain.
In addition, we conduct experiments on several real-world datasets to demonstrate
that KM-GAN is really capable to generate realistic and diverse images without mode collapse.
In the future, we would further pay attention to proving the positive correlation between high-quality synthesized
images and high clustering accuracy and utilize the relationship to improve performance of both tasks.

\section*{Acknowledgments}
The authors would like to thank Jiaxiang Guo, Tianli Liao, Yifang Xu,
Bowen Wu, Mengya Zhang, Chengdong Zhao and Dong Wang for their helpful advices.

\section*{References}

\bibliography{mybibfile}

\linenumbers

\end{document}